\documentclass[twoside,leqno,twocolumn]{article}

\usepackage[letterpaper]{geometry}
\usepackage{soul}
\usepackage{ltexpprt}
\usepackage{hyperref}
\usepackage{xcolor}
\usepackage{graphicx}
\usepackage{subcaption}
\usepackage{tablefootnote}
\usepackage{threeparttable}
\usepackage{amsmath,amssymb,amsfonts}
\usepackage{multirow}
\usepackage{tabularx}
\usepackage{comment}
\usepackage{url}

\providecommand{\keywords}[1]
{
  \small	
  \textbf{\textit{Keywords---}} #1
}
\newcommand{\github}[1]{%
   \href{#1}{\faGithubSquare}%
}

\begin{document}

\newcommand\relatedversion{}

\title{\Large DeepGLSTM: Deep Graph Convolutional Network and LSTM based approach for predicting drug-target binding affinity\relatedversion}

\author{Shrimon Mukherjee  \thanks{Indian Association for the Cultivation of Science.} \thanks{Equal Contribution to this work} \\  \href{iacsshrimon@gmail.com}{iacsshrimon@gmail.com} 
\and Madhusudan Ghosh \footnotemark[1] \footnotemark[2] \\  \href{madhusuda.iacs@gmail.com}{madhusuda.iacs@gmail.com} \and Partha Basuchowdhuri \footnotemark[1] \\ \href{partha.basuchowdhuri@iacs.res.in}{partha.basuchowdhuri@iacs.res.in}}
%\address{{$\dagger$} The authors equally contributed to this paper.}
\date{}

\maketitle

% Copyright Statement
% When submitting your final paper to a SIAM proceedings, it is requested that you include
% the appropriate copyright in the footer of the paper.  The copyright added should be
% consistent with the copyright selected on the copyright form submitted with the paper.
% Please note that "20XX" should be changed to the year of the meeting.

% Default Copyright Statement
\fancyfoot[R]{\scriptsize{Copyright \textcopyright\ 2022 by SIAM\\
Unauthorized reproduction of this article is prohibited}}

% Depending on which copyright you agree to when you sign the copyright form, the copyright
% can be changed to one of the following after commenting out the default copyright statement
% above.

%\fancyfoot[R]{\scriptsize{Copyright \textcopyright\ 20XX\\
%Copyright for this paper is retained by authors}}

%\fancyfoot[R]{\scriptsize{Copyright \textcopyright\ 20XX\\
%Copyright retained by principal author's organization}}

\pagenumbering{arabic}

\begin{abstract} \small\baselineskip=9pt Development of new drugs is an expensive  and time-consuming process. Due to the world-wide SARS-CoV-2 outbreak, it is essential that new drugs for SARS-CoV-2 are developed as soon as possible. Drug repurposing techniques can reduce the time span needed to develop new drugs by probing the list of existing FDA-approved drugs and their properties to reuse them for combating the new disease. We propose a novel architecture DeepGLSTM, which is a Graph Convolutional network and LSTM based method that predicts binding affinity values  between the FDA-approved drugs and the viral proteins of SARS-CoV-2. Our proposed model has been trained on Davis, KIBA (Kinase Inhibitor Bioactivity), DTC (Drug Target Commons), Metz, ToxCast and STITCH datasets. We use our novel architecture to predict a Combined Score (calculated using Davis and KIBA score) of 2,304 FDA-approved drugs against 5 viral proteins. On the basis of the Combined Score, we prepare a list of the top-18 drugs with the highest binding affinity for 5 viral proteins present in SARS-CoV-2. Subsequently, this list may be used for the creation of new useful drugs.\end{abstract}

\section{Introduction}
The discovery of drugs, by the traditional approach, is time-consuming and expensive. Generally, it costs billions of US dollars and takes about 10-15 years for a drug to be accepted or rejected by FDA (US Food and Drug Administration) \cite{ashburn2004drug,roses2008pharmacogenetics}. The traditional approach of drug discovery goes under several phases of trials \cite{ashburn2004drug}. To reduce time and cost, we use already approved drugs to identify their use in combating SARS-CoV-2. This technique of discovering new applications of drugs is known as drug repurposing or repositioning \cite{strittmatter2014overcoming}.\\
A high throughput screening experiment examines bioactivities between drugs and proteins in this traditional approach, which makes it a costly and time-consuming process \cite{cohen2002protein,noble2004protein}. This is an infeasible task  as there are millions of drug-like compounds \cite{deshpande2005frequent} and hundreds of potential targets. As an example, 500 protein kinases \cite{manning2002protein} are responsible for  modification of about 30\% of human proteins \cite{stachel2014maximizing}.
Protein plays an important role in drug-target interaction prediction. Proteins are the most important components for most of the functions within and outside  human cells. The three-dimensional structure of a protein and its spatial orientation determines the role of a protein. Therefore,  structural changes in a protein can significantly alter its functionality~\cite{alberts2002shape}. In drug discovery, many drugs are designed to bind specific proteins. Any structural change of drugs may thus alter their properties to bind with the proteins. As a result, an important problem in computational drug discovery is to predict whether a drug can bind with a specific protein or not. This problem is popularly known as drug-target interaction prediction, and has become a topic of significant research interest in recent years.
We use drug-target interaction prediction in order to reuse FDA-approved drugs on new target proteins in the repurposing or repostioning \cite{strittmatter2014overcoming} technique of drug discovery. In DTI (drug-target interaction) \cite{ozturk2018deepdta}, the goal is to predict the binding affinity between the FDA-approved drugs and the new target proteins. Machine learning methods may be used for predicting the drug-target binding affinity.

Drug-target interaction is measured in terms of binding affinity \cite{ozturk2018deepdta}. Greater interaction between the drug-target pairs indicate a stronger readout for binding affinity. It is evaluated in terms of \textit{inhibition constant} ($K_{i}$), \textit{dissociation constant} ($K_{d}$), changes in \textit{free energy} measures ($\delta$G, $\delta$H), the \textit{half-maximal inhibitory constant} ($IC_{50}$) \cite{wang2020deep}, \textit{half-maximal activity concentration} ($AC_{50}$)~\cite{feng2018padme}, KIBA score~\cite{tang2014making} and SCORES (used in STITCH database)~\cite{kuhn2007stitch}. In~\cite{tang2014making}, KIBA scores were calculated to optimize the consistency between $K_{i}$, $K_{d}$ and $IC_{50}$ by making use of the available statistical information. Here, we introduce, C\textsubscript{b}, a Combined Score (by combining both $pK_{d}$ and KIBA scores, as mentioned in Eqn.~\ref{comb}) to predict binding affinity between FDA-approved drugs and new targets for SARS-CoV-2.

Many statistical and machine learning models \cite{kinnings2011machine,iskar2012drug,corsello2017drug} are used in the repurposing or repositioning approach of the discovery of drugs, i.e., to predict DTI (drug-target interaction) of FDA-approved drugs on new targets. The machine learning techniques, which have been previously used for DTI prediction, are affinity similarity (SimBoost) \cite{he2017simboost} and Kronecker regularized least squares based (KronRLS) \cite{cichonska2017computational,cichonska2018learning}.\par
SimBoost  \cite{he2017simboost} is a gradient boosting method that depends on the features extracted from the drugs, the targets, and the drug-target pairs. This approach uses feature engineering to build three types of features:
\begin{itemize}
    \item object-based features.
    \item network-based features collected from a homogeneous network - such as neighborhood  statistics, network metrics (betweenness, closeness, etc.), and PageRank scores collected from drug-drug, target-target networks.
    \item network-based features collected from a heterogeneous network - such as drug-target networks.
\end{itemize}\par
These features are supplied into a supervised learning method named gradient boosting regression trees, derived from the gradient boosting machine learning model.

In the KronRLS \cite{cichonska2017computational,cichonska2018learning} approach, kernels from drugs and targets are built from their molecular descriptors, fed into a regularized least squares regression model (RLS) to predict the binding affinity.
In recent years, DTI predictions have shifted from machine learning models to deep learning models \cite{ozturk2018deepdta}. Deep learning models predict better than the above-mentioned machine learning models. Some of the deep learning models proposed earlier for drug repurposing, are DeepDTA~\cite{ozturk2018deepdta}, WideDTA~\cite{ozturk2019widedta}, MT-DTI~\cite{shin2019self}, DeepCPI~\cite{tsubaki2019compound}, GANsDTA~\cite{zhao2020gansdta}, Attention-DTA~\cite{8983125}, 1-D CNN~\cite{majumdar2021deep}, DeepGS~\cite{lin2020deepgs} and  GraphDTA~\cite{nguyen2021graphdta}.\\
In DeepDTA \cite{ozturk2018deepdta}, drugs (provided in SMILES notation) and targets are represented as sequence of characters. In WideDTA~\cite{ozturk2019widedta}, drugs and proteins are represented as words instead of characters as in DeepDTA~\cite{ozturk2018deepdta}. Similarly, other deep learning based models are DeepGS ~\cite{lin2020deepgs}, DeepCPI ~\cite{tsubaki2019compound}, MT-DTI ~\cite{shin2019self}, Attention-DTA~\cite{8983125} and 1-D CNN~\cite{majumdar2021deep}, which are also used for drug-target interaction (DTI) prediction.

Earlier, Graph Neural Network (GNN) based works ~\cite{lin2020deepgs}~\cite{nguyen2021graphdta} did not capture proper topological information of the required graph (i.e., generated from the SMILES notation of any drug) by providing simple adjacency representation as input to the GNN module. However, ~\cite{lin2020deepgs}~\cite{nguyen2021graphdta} used 1-D CNN to encode protein sequence information. It is well-known from existing literature that LSTM performs better than 1-D CNN to capture sequential information~\cite{yin2017comparative} and therefore, has been used in our model.\\
Our proposed model \textbf{DeepGLSTM} introduces a novel Graph Convolution Network (GCN) block to pass the drug compound information using power graph representation to capture graph topological information to achieve state-of-the-art (SOTA) results in neural drug repurposing domain.

\begin{comment}
\textcolor{red}{We clearly see that such models (i.e., ~\cite{ozturk2018deepdta},~\cite{zhao2020gansdta}, etc.) suffer for incorporating topological information from the drugs because 1-D CNN does not incorporate topological information from SMILES and for target they suffers for getting long sequential formation using 1-D CNN. These are the drawbacks for the models (~\cite{ozturk2018deepdta},~\cite{zhao2020gansdta} etc). We use GCN~\cite{kipf2016semi} to incorporate topological information from drugs and to capture the sequential information we use LSTM, because LSTM performs better than 1 D CNN to capture the sequential information~\cite{yin2017comparative}. The GCN based models like ~\cite{nguyen2021graphdta} uses three GCN layers but they fails to capture all the topological information from the drugs. On the other our model captures more topological information because our model produces better accuracy than the GCN based models.Thus the GCN based models has drawbacks.} 
\end{comment}

Our work makes the following contributions:
\begin{itemize}
    \item \textbf{DeepGLSTM} introduces a Graph Convolution Network (GCN) block to process the drug compounds using power graph representation and a   bidirectional-LSTM (Bi-LSTM) layer to process the protein sequence. The experimental results show that our model outperforms the previous state-of-the-art results for the affinity prediction task.
    \item We train our model on multiple benchmark datasets, Davis~\cite{davis2011comprehensive}, KIBA~\cite{tang2014making},
    DTC~\cite{tang2018drug}, Metz~\cite{metz2011navigating}, ToxCast~\cite{Toxcast} and STITCH~\cite{kuhn2007stitch}. We also apply our proposed model to predict the affinity score (C\textsubscript{b} Score, Eqn.~\ref{comb}) between FDA-approved drugs and the 5 viral proteins~\cite{zhou2020addendum,zhang2020protein,walls2020structure,wrapp2020cryo,gao2020structure} of SARS-CoV-2. 
\end{itemize}

\section{Related Work}
Drug-target affinity predictions are presently performed in two ways. The first one is simply by using traditional machine learning based methods ~\cite{kinnings2011machine,iskar2012drug,corsello2017drug}. The second one consists of adoption of deep learning based methods~\cite{ozturk2018deepdta}. Nowadays, prediction of drug-target binding affinity has shifted from traditional machine learning models to deep learning models and most of the recent works are based on neural networks, i.e., on deep learning based approaches.

%\cite{he2017simboost}\st{ used a gradient based approach for predicting drug-target affinity values. Similarly},~\cite{cichonska2017computational,cichonska2018learning} \st{used Kronecker regularized least squares based approach for predicting drug-target binding affinity. These two methods are traditional machine learning based approach and they heavily depend on the nature of the feature engineering techniques adopted.}

\cite{tsubaki2019compound}
used an end-to-end deep learning approach for predicting drug-target affinity. Similarly,~\cite{ozturk2018deepdta} introduced CNN based approach to predict affinity between drugs and targets, and this trend continued in ~\cite{ozturk2019widedta} and~\cite{majumdar2021deep}. Similarly,~\cite{shin2019self} gives a self-attention based approach for predicting affinity.~\cite{zhao2020gansdta} proposed a generative adversarial network (GAN) to learn useful patterns within labeled and unlabeled sequences and takes advantage of convolutional regression to predict affinity. In this method, two partial GANs, one for the feature extraction from the raw protein sequences, and another for the SMILES strings, were used with a CNN based regression network, for the purpose of affinity prediction. But~\cite{zhao2020gansdta} did not discuss about how GAN would perform when trained on small datasets. Again,~\cite{8983125} used CNN architecture followed by attention based mechanisms to determine the significance of the SMILES sequences of the drugs for interactions with proteins, thereby using it for affinity score prediction. Nevertheless, all of these methods convert drug compounds into corresponding string representations, which are not ideal to represent drug molecules. Using such strings may lead to exclusion of the topological information of the drug molecules. As a result, it decreases the performance of the above-discussed method and its power of predicting drug-target affinity.~\cite{abbasi2020deepcda} used both LSTM and CNN for predicting drug-target binding affinity score.~\cite{shim2021prediction} introduced a similarity-based method using CNN to perform the required task.~\cite{govinda2019kinasepkipred} used DTC as well as Metz dataset to perform the required task.~\cite{feng2018padme} used ToxCast dataset to predict drug-target binding affinity score.

In contrast to the aforementioned  models,~\cite{nguyen2021graphdta} used topological information of drug molecules using graph neural networks. In this method, chemical structures of drugs (provided in SMILES notation) are represented as graphs, where the nodes are the atoms, the edges represent interactions between atoms, and the proteins are represented as strings made of characters. Specifically, graph convolutional networks (GCN), graph attention networks (GAT) and graph isomorphism networks (GIN) were used to predict drug-target affinity. Similarly,~\cite{lin2020deepgs} used a combination of GAT and 1-D CNN for predicting affinity between drugs and target.

We compare our results with KronRLS~\cite{cichonska2017computational,cichonska2018learning}, SimBoost~\cite{he2017simboost}, DeepDTA~\cite{ozturk2018deepdta}, MT-DTI~\cite{shin2019self}, DeepCPI~\cite{tsubaki2019compound}, WideDTA~\cite{ozturk2019widedta}, GANsDTA~\cite{zhao2020gansdta}, AttentionDTA~\cite{8983125}, 1D-CNN~\cite{majumdar2021deep}, DeepGS~\cite{lin2020deepgs} and GraphDTA~\cite{nguyen2021graphdta}.

\section{Methodologies}
\subsection{Proposed Architecture}
\begin{figure}[!t]
\hspace{-1.7cm}
\includegraphics[scale=0.65]{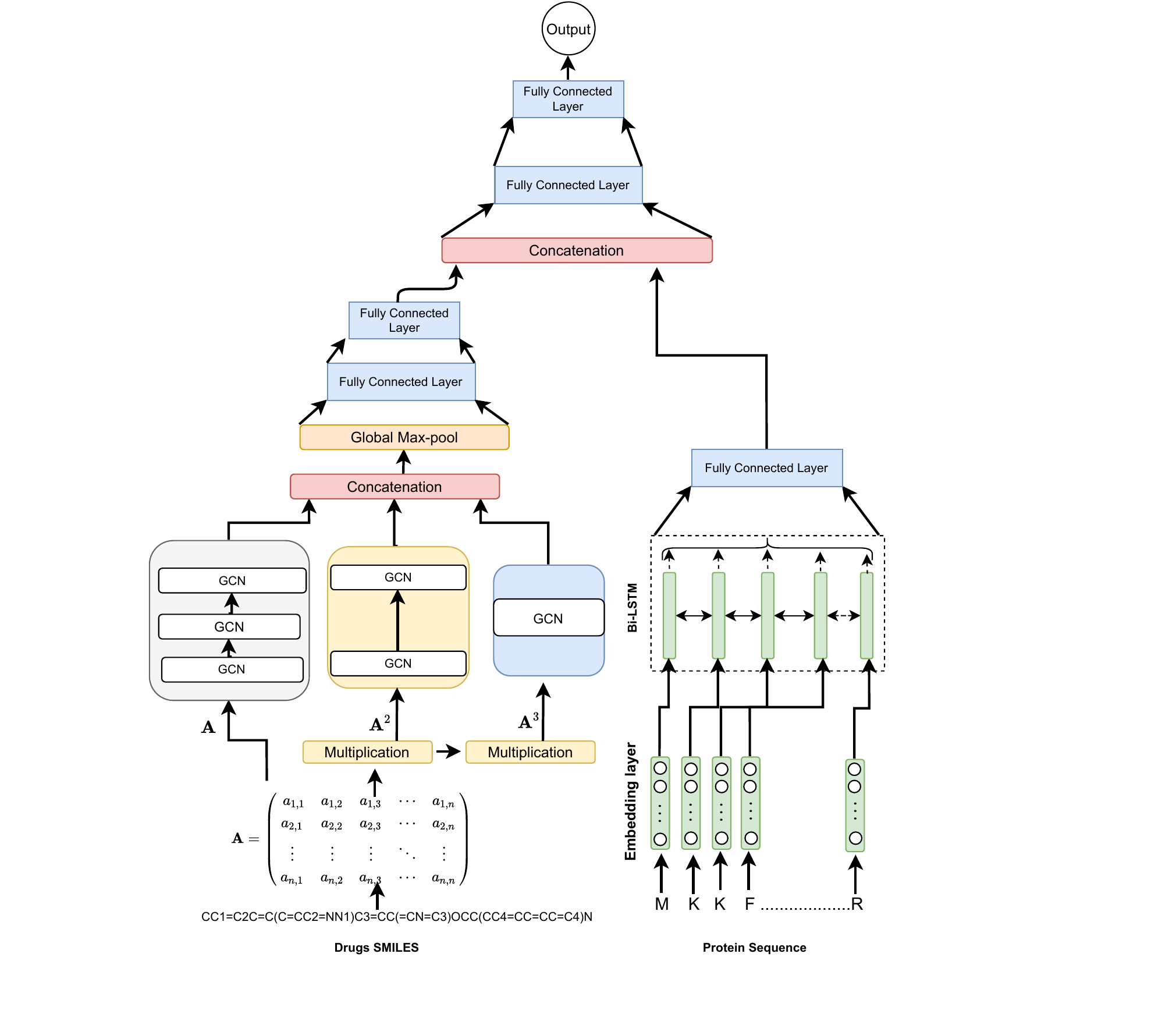}
\caption{Model architecture of \textbf{DeepGLSTM}}
\label{fig:DTA}
\end{figure}

In this section, we present our novel architecture 
\textbf{DeepGLSTM} to predict the required affinity score for drug-target binding. The architecture of our proposed model is shown in Fig. \ref{fig:DTA}. \textbf{DeepGLSTM} has two functional modules. The first one helps to capture the topological information from the drug molecules. On the other hand, the second one captures the sequential information from the targeted protein structures.
\\

In most of the previous works, drugs and proteins have been encoded using one-hot representation. Later, ~GraphDTA~\cite{nguyen2021graphdta}, ~DeepGS~\cite{lin2020deepgs} used graph representation to embed the drug compounds and one-hot representation for protein structures.
Experimental results show that providing graph representation as an input improves the model performance and produces better results. We follow this technique to embed the drug compound topological information and feed it as an input.
Similarly, for protein sequences, one could also represent them as graphs. But, doing so is more difficult because the tertiary structure of a protein is not always available in a reliable form. So, for protein sequences we use the popular one-hot encoding representation~\cite{nguyen2021graphdta}.\\ We represent each drug compound by its corresponding SMILES (Simplified Molecular Input Line Entry System)\cite{weininger1988smiles} notation. This notation always considers the compounds as a graph of the interactions between atoms (i.e., the nodes). To describe a node feature, we use a set of atomic features adapted from DeepChem \cite{ramsundar2019deep}. Here, each node is represented as a multi-dimensional binary feature vector. The feature vector of every node demonstrates five pieces of information (i.e., the atom symbol, the number of adjacent atoms, the number of adjacent hydrogens, the implicit valence of the atom and whether the atom is in an aromatic structure)~\cite{ramsundar2019deep}. Then we use the open-source chemical informatics software RDKit \cite{landrum2013rdkit} to convert the SMILES data to its corresponding molecular graph representation (i.e., adjacency representation $A\in R^{N\times N}$ ), which helps us to extract the required atomic features.

The protein sequence is a string of ASCII characters, which represents amino acids. We initialize the protein sequence embedding layer by mapping every amino acid to an English letter and subsequently, by marking the amino acid with the sequential index of that letter in the English alphabet. We pass the embedding representation ($c\in\mathbb R^{d_p}$, where $d_p$ is the dimension of the protein sequence embeddings) to a Bi-LSTM layer, which captures the dependencies between the characters in a sequence $(p=[\mathrm{c_1,c_2...c_n}])$ of length $n$. We get the output representation $h_t \in\mathbb R^{2d_l}$, where $d_l$ represents the number of output units used in each LSTM cell. We compute Eqn. \ref{LSTM} for execution of the LSTM.

\begin{equation}
\label{LSTM}
\begin{split}
    \overrightarrow{h_{t}} &= \overrightarrow{LSTM}(c_t,h_{t-1})\\
    \overleftarrow{h_{t}} &= \overleftarrow{LSTM}(c_{t},h_{t+1})\\
    h_{t} &= \overrightarrow{h_{t}}||\overleftarrow{h_{t}}
\end{split}
\end{equation}

\begin{table}[h!]
    \footnotesize
    \centering
    \scalebox{0.8}{\begin{tabular}{|c|c|c|c|}
    \hline
     Dataset & \# of drugs & \# of targets & Total number of\\
      &       (compounds) &  proteins & drug-target pairs used\\
      \hline
      Davis($k_{d}$) & 68 & 442 & 30056\\
      \hline
      KIBA & 2111 & 229 & 118254\\
      \hline
      DTC($pk_{i}$) & 5983 & 118 & 67894\\
      \hline
      Metz($pk_{i}$) & 1471 & 170 & 35307\\
      \hline
      ToxCast($AC_{50}$) & 7657 & 328 & 342869\\
      \hline
      STITCH & 724471 & 15258 & 1244420\\
      \hline
    \end{tabular}
    }
    \caption{Dataset statistics}
    \label{tab:Datasets}
\end{table}

To predict the affinity score of drug-protein interaction, understanding the interaction of each node with its neighboring nodes is very important. Generally, adjacency representation of any graph holds the connectedness relationship of any node with the other ones, which are connected to that node by an edge. So, to represent the graph level features for every drug compound, incorporation of multi-hop connectedness relationship between the nodes present in a graph may turn out to be important. For this reason, we introduce the idea of using GCN in our architecture. It is a well-known fact that a randomly initialized GCN is strong enough to produce important feature representations, which can capture the connectedness relationship between the graph nodes.
We introduce three blocks consisting of GCNs. In the first block, we stack three GCN layers. In the
second block, we stack two GCN layers and in the final block we use only one GCN layer. For every drug compound, we compute a simple propagation rule as mentioned in \cite{kipf2016semi} by taking the adjacency representation ($A\in R^{N\times N}$) generated from the RDKit tool and a node feature matrix $X\in R^{N\times C}$ (where $C$ is the number of features per node) as inputs in the first block. To overcome the degree normalization problem of the adjacency representation, we compute the normalized adjacency representation ($A_{norm}$) as in Eqn. \ref{gcn1}, where $D\in R^{N\times N}$ is the degree matrix representation of $A$.

\begin{comment}

\begin{table}[h!]
    \footnotesize
    \begin{center}
    \scalebox{0.9}{ \begin{tabular}{r|llll}
        \hline
             & \textbf{Proteins} & \textbf{Compounds} & \textbf{Interactions} \\
        \hline
        \multirow{1}{*}{Davis($k_{d}$)}
        & 442 & 68 & 30056 \\
        \hline
        \multirow{1}{*}{KIBA}
        & 229 & 2111 & 118254 \\
        \hline
        \multirow{1}{*}{DTC($pk_{i}$)} & 118 & 5983 & 67894\\
        \hline
        \multirow{1}{*}{Metz($pk_{i}$)} &  & - & -\\
        \hline
        \multirow{1}{*}{Toxcast($AC_{50}$)} & - & - & -\\
        \hline
        \multirow{1}{*}{Stitch (Combined Score)} & - & - & -\\
        \hline
        \end{tabular}}
    \end{center}
    \vspace{-2mm}
    \caption{Dataset statistics}
    \label{tab:Datasets}
\end{table}
\end{comment}
\begin{equation}
    \label{gcn1}
    \begin{split}
        A_{norm} = D^{-1/2}AD^{-1/2}
    \end{split}
\end{equation}
The Eqn. \ref{gcn2}  makes the first block workable, where $W$ is the trainable weight, $H_{e}^{0}=X$ is the $i^{th}$ layer output representation and $\sigma$ is a non-linear activation function. Thus, $i^{th}$ layer of the GCN module produces a global representation ($H_{e}^{i} \in R^{N\times M}$) for $A$.
\begin{equation}
    \label{gcn2}
    \begin{split}
        H_{e}^{i} = \sigma(A_{norm}H_{e}^{(i-1)}W^{(i-1)})
    \end{split}
\end{equation}

In a graph, every node \textit{v} is connected to every node \textit{u} that belongs to its neighborhood $\Gamma$(\textit{v}) by an edge. Every node \textit{w} in $\Gamma$(\textit{u}) that are not present in $\Gamma$(\textit{v}) has a  shortest-path distance of 2 from \textit{v}. In other words, they are two hops away from \textit{v}. If \textit{v} is connected to all such nodes that are two hops away from it, we get a graph that can be represented by using $A^{2}$. Such a graph can be referred to as a power graph of power 2. Similarly, to further increase the local reachability of \textit{v}, the value of the exponent of the power graph may be increased. Note that, increasing the exponent of the power graph would typically lead to much denser graphs due to its increased reachability.

In the GCN second block, we take squared representation ($A^{2}\in R^{N\times N}$) of $A\in R^{N\times N}$ and the same node feature representation $X$ as input. By similar argument, we compute the normalized adjacency representation $A^{2}_{norm}$ as in Eqn. \ref{gcn3}, where $D^{\prime}$ is the degree representation of $A^{2}$. 
\begin{equation}
    \label{gcn3}
    \begin{split}
        A^{2}_{norm} = D^{\prime ^{-1/2}}A^{2}D^{\prime ^{-1/2}}
    \end{split}
\end{equation}
Eqn. \ref{gcn4} makes the second GCN block workable, $W^{\prime}$ is the trainable weight, $H^{\prime ^{0}}_{e} = X$ is the $i^{th}$ layer output representation and $\sigma$ is a non-linear activation function. Thus, this GCN block produces a global representation ($H_{e}^{\prime ^ {i}}\in {R^{N\times M^{\prime }}}$) for $A^{2}$. 
\begin{equation}
    \label{gcn4}
    \begin{split}
        H_{e}^{\prime^{i}} = \sigma(A^{2}_{norm}H_{e}^{\prime ^ {(i-1)}}W^{\prime ^{(i-1)}})
    \end{split}
\end{equation}
Finally, GCN layer of the third block takes cubic representation ($A^{3}\in R^{N\times N}$) of $A\in R^{N\times N}$ and the same feature $X$ as input. By similar argument, we compute the normalized adjacency representation $A^{3}_{norm}$ as in Eqn. \ref{gcn5}, where $D^{\prime\prime}$ is the degree representation of $A^{3}$.

\begin{equation}
    \label{gcn5}
    \begin{split}
        A^{3}_{norm} = D^{\prime \prime ^{-1/2}}A^{3}D^{\prime \prime ^{-1/2}}
    \end{split}
\end{equation}
Similarly, Eqn. \ref{gcn6} makes the third GCN block workable and the parameters are same as discussed above. This block produces the required global representation ($H_{e}^{\prime \prime ^{i}}\in R^{N\times M^{\prime \prime}}$) for $A^{3}$. 
\begin{equation}
    \label{gcn6}
    \begin{split}
        H_{e}^{\prime \prime ^{i}} = \sigma(A^{3}_{norm}H_{e}^{\prime \prime^ {(i-1)}}W^{\prime \prime^{(i-1)}})
    \end{split}
\end{equation}

We concatenate the output representations of the three blocks as shown in Eqn. \ref{gcn7} to get the final graph level representation for every drug compound. 
\begin{equation}
    \label{gcn7}
    \begin{split}
        H_{e} = (H_{e}^{i}.H_{e}^{\prime^{i}}.H_{e}^{\prime \prime ^{i}})
    \end{split}
\end{equation}\par
Subsequently, we
pass $H_{e}$ into a global max-pool layer followed by some fully connected layers to get the representation, which makes the computation fast and efficient. Then we concatenate the output representations of the Bi-LSTM and the fully connected layers. Finally, we feed the concatenated representation into some fully connected layers  followed by the output layer to predict the affinity score.

\subsection{Baselines}
\label{baseline}
We consider the following models as baselines - traditional machine learning based methods KronRLS~\cite{cichonska2017computational,cichonska2018learning}, SimBoost~\cite{he2017simboost}, deep learning based models DeepDTA~\cite{ozturk2018deepdta}, WideDTA~\cite{ozturk2019widedta}, GANsDTA~\cite{zhao2020gansdta}, AttentionDTA~\cite{8983125}, DeepCPI~\cite{tsubaki2019compound}, MT-DTI~\cite{shin2019self}, GraphDTA~\cite{nguyen2021graphdta} and DeepGS~\cite{lin2020deepgs}.

\subsection{Dataset}
\label{data}
Here, we have used Davis~\cite{davis2011comprehensive}, KIBA~\cite{tang2014making}, DTC~\cite{tang2018drug}, Metz~\cite{metz2011navigating}, ToxCast~\cite{Toxcast} and
STITCH\footnote[1]{\url{http://stitch.embl.de/}}~\cite{kuhn2007stitch} datasets for our experiments as summarized in Table \ref{tab:Datasets}.\par
In our experiments, we transformed the $K_{d}$ value of the Davis dataset into log space  $pK_{d}$, similar to \cite{he2017simboost}, by using Eqn.~\ref{logs}. We also divide SCORES metric for STITCH dataset by 100 to make the training process simpler.

\begin{equation}
    \label{logs}
    \begin{split}
        pK_{d} = - \log_{10}(\frac{K_{d}}{1e9})
    \end{split}
\end{equation}

\subsection{Parameter Settings}
In our experiments, we fixed the length of the protein sequence ($n$) to 1000 for all the datasets. The dimension ($2d_i$) of the hidden vector representation of the Bi-LSTM layer is of 128. We also fixed the number of input node features ($C$) of the required drug compounds to 78. We set the final hidden vector representation ($M$) of the first GCN block to 312. Similarly, we also set the output representations ($M^{\prime}$, $M^{\prime \prime}$) for the rest of the GCN blocks to 156, 78 respectively. Throughout our experiment, we have considered a dropout probability rate ($p$) of 0.2 and we used ADAM~\cite{kingma2015adam} optimizer with learning rate 0.0005. Since this is a regression problem, we use the mean squared error (\textbf{MSE}) as the loss function to evaluate the training of our model. For all the experiments, we train our model to 1000 epochs. We use batch size ($B$) of 512 for KIBA, DTC, Metz, ToxCast and STITCH and batch size of 128 for Davis dataset. We followed the parameter settings of GraphDTA~\cite{nguyen2021graphdta} to carry out baseline experiments for the mentioned datasets (cf. section \ref{data}). Reported results of the baselines models in Table~\ref{tab:Model_per1} were taken from the previous literature works. Note that the baseline results of Table~\ref{tab:Model_per2} were obtained by carrying out the required experiments using the source code given in the corresponding work~\cite{nguyen2021graphdta}. We used Google Colab PRO plus to carry out the experiments.

\begin{equation}
    \label{MSE}
    \begin{split}
        MSE = \frac{1}{B}\sum_{i=1}^{B}(P_{i}-Y_{i})^{2}
    \end{split}
\end{equation}

\begin{equation}
    \label{CI}
    \begin{split}
        CI = \frac{1}{Z}\sum_{\delta_{i}>\delta_{j}}h(b_{i}-b_{j})\\
h(x) = \begin{cases} 1, &\text{if }x \ge 0 \\ 0.5, &\text{if }x = 0
\\0, &\text{if } x \leq 0 
\end{cases}    
    \end{split}
\end{equation}

\begin{equation}
\label{rm2}
    \begin{split}
        r_{m}^{2} = r^{2}\times (1- \sqrt{r^{2}-r_{0}^{2}}) 
    \end{split}
\end{equation}

\begin{equation}
    \label{comb}
       \begin{split}
          KB = \text{KIBA}\\
          DB = \text{Davis ($pK_{d}$)}\\
          K = 1 - \frac{KB_{i}}{\max_{i}(KB)}\\
          D = \frac{DB_{i}}{\max_{i}({DB})}\\
          \text{C\textsubscript{b}} = \frac{1}{2}\sum_{i=1}^{n'}(K_{i} + D_{i})
       \end{split}
\end{equation}

\subsection{Results}
\label{res}
We use (\textbf{MSE}), concordance index(\textbf{CI}) and $\mathbf{r_{m}^{2}}$ \textbf{index} to evaluate the performance of our model.

\begin{table}[h]
\footnotesize
\begin{center}
\scalebox{0.7}{   \begin{tabular}{|c|c|c|c|c|c|c|}
        \hline
            {}  &\multicolumn{3}{|c|}{{\bf Davis ($\mathbf{pK_{d}}$)}} & \multicolumn{3}{|c|}{{\bf KIBA}}\\
        \hline
        {\bf Model} & {\bf MSE} & {\bf CI} & $\mathbf{r_{m}^{2}}$ & {\bf MSE} & {\bf CI} & $\mathbf{r_{m}^{2}}$\\
        \hline
        KronRLS \cite{cichonska2017computational,cichonska2018learning} & 0.379 & 0.869 & 0.407 &  0.411 & 0.782 & 0.342\\
        \hline
        SimBoost \cite{he2017simboost} & 0.282 & 0.873 & 0.644 & 0.222 & 0.836 & 0.629\\
        \hline
        DeepDTA \cite{ozturk2018deepdta} & 0.261 & 0.878 & 0.630 &  0.194 & 0.863 & 0.673\\
        \hline
        MT-DTI (Wo-FT) \cite{shin2019self} & 0.268 & 0.875 & 0.633 & 0.220 & 0.844 & 0.584\\
        \hline
        MT-DTI \cite{shin2019self} &  0.245 & 0.887 & 0.665 & 0.193 & 0.882 & 0.738\\
        \hline
        DeepCPI \cite{tsubaki2019compound}& 0.293 & 0.867 & 0.607 &  0.211 & 0.852 & 0.657\\
        \hline
        WideDTA \cite{ozturk2019widedta} & 0.262 & 0.886 & 0.633 & 0.179 & 0.875 & 0.675\\
        \hline
        GANsDTA \cite{zhao2020gansdta} &
        0.276 & 0.881 & 0.653 & 0.224 & 0.866 & 0.775\\
        \hline
        Attention-DTA \cite{8983125} & 0.245 & 0.887 & 0.657 & 0.162 & 0.882 & 0.735\\
        \hline
        1D-CNN \cite{majumdar2021deep} & - & - & - & 0.700 & - & - \\
        \hline
        DeepGS \cite{lin2020deepgs} & 0.252 & 0.880 & {\bf 0.686} & 0.193 & 0.860 & 0.684\\
        \hline
        GraphDTA (GCN) \cite{nguyen2021graphdta} & 0.254 &  0.880 &- & 0.139 & 0.889 & -\\
        \hline
        GraphDTA (GAT\_GCN) \cite{nguyen2021graphdta}  & 0.245 & 0.881 & - & 0.139 & 0.891 & -\\
        \hline
        GraphDTA (GAT) \cite{nguyen2021graphdta}
        & 0.232 & 0.892 & 0.662 & 0.179 & 0.866 & 0.671\\
        \hline
        GraphDTA (GIN) \cite{nguyen2021graphdta} & {\bf 0.229} & 0.893 & 0.649 & 0.147 & 0.882 & 0.684\\
        \hline
        {\bf DeepGLSTM}\tablefootnote[1]{Open-source code is available at \url{https://github.com/MLlab4CS/DeepGLSTM.git}} & 0.232 & {\bf 0.895} & 0.680 & {\bf 0.133} & {\bf 0.897} &  {\bf 0.792}\\
        \hline
        \end{tabular}
        }
    \end{center}
    %\vspace{-2mm}
    \caption{Performance comparison of \textbf{DeepGLSTM} against all other baseline models (cf. section~\ref{baseline}) on Davis ($pK_{d}$) and KIBA.}
    \label{tab:Model_per1} 
\end{table}

\begin{table*}[!t]
\footnotesize
\begin{threeparttable}[b]
\centering{
\scalebox{0.9}{\begin{tabular}{|c|c|c|c|c|c|c|c|c|c|c|c|c|c|}
        \hline
            {}  &\multicolumn{3}{|c|}{{\bf DTC ($\mathbf{pK_{i}}$)}} & \multicolumn{3}{|c|}{{\bf Metz ($\mathbf{pK_{i}}$)}} & \multicolumn{3}{|c|}{\bf Toxcast ($\mathbf{AC_{50}}$)} & \multicolumn{2}{|c|}{\bf Stitch}\\
            & \multicolumn{3}{|c|}{} & \multicolumn{3}{|c|}{} & \multicolumn{3}{|c|}{} & \multicolumn{2}{|c|}{\bf (SCORES)}\\
        \hline
        {\bf Model} & {\bf MSE} & {\bf CI} & $\mathbf{r_{m}^{2}}$ & {\bf MSE} & {\bf CI} & $\mathbf{r_{m}^{2}}$ & {\bf MSE} & {\bf CI} & $\mathbf{r_{m}^{2}}$ & {\bf MSE} & $\mathbf{r_{m}^{2}}$\\
        \hline
        GraphDTA (GCN)~\cite{nguyen2021graphdta} & 0.317 & 0.878 & 0.812 & 0.317
        & 0.801 & 0.620 & 0.315 & 0.917 & 0.561 & 1.022 & 0.424 \\
        \hline
        GraphDTA (GAT\_GCN)~\cite{nguyen2021graphdta}  & 0.200 & 0.857 & 0.790 & 0.333 & 0.795 & 0.602 & 0.317 & 0.917 & {\bf 0.566} & - & -\\
        \hline
        GraphDTA (GAT)~\cite{nguyen2021graphdta} & 0.195 & 0.859 & 0.788 & 0.393 & 0.775 & 0.549 & 0.346 & 0.911 & 0.535 & - & - \\
        \hline
        GraphDTA (GIN)~\cite{nguyen2021graphdta} & 0.176 & 0.876 & 0.798 & 0.317 & 0.800 & {\bf 0.645} & 0.324 & 0.915 & 0.555 & - & -\\
        \hline
        {\bf DeepGLSTM} & {\bf 0.149} & {\bf 0.895} & {\bf 0.841} & {\bf 0.294} & {\bf 0.810} & 0.640 & {\bf 0.313} & {\bf 0.919} & 0.565 & \bf{0.992} & \bf{0.430}\\
        \hline
        \end{tabular}
        }
        }
    %\vspace{-2mm}
    \caption{Performance comparison between~\bf DeepGLSTM and~\bf GraphDTA on {\bf DTC ($\mathbf{pK_{i}}$)},~{\bf Metz ($\mathbf{pK_{i}}$)},~{\bf Toxcast ($\mathbf{pK_{i}}$)} and~{\bf Stitch (SCORES)}.}
    \label{tab:Model_per2}
    \end{threeparttable}
\end{table*}

\begin{comment}

\begin{figure}[!t]
    \centering
    \includegraphics[scale=0.4]{Davis_cross_valid_new.pdf}
    \caption{Predictions from our model (\textbf{DeepGLSTM}) versus measured (real) binding affinity values for the Davis dataset ($pK_{d}$). }
    \label{fig:davis_cross}
\end{figure}

\begin{figure}[!t]
    \centering
    \includegraphics[width=7cm,height=5cm]{kiba_cross_valid_new.pdf}
    \caption{Predictions from our model (\textbf{DeepGLSTM}) versus measured (real) binding affinity values for the KIBA dataset (KIBA score). }
    \label{fig:kiba_cross}
\end{figure}

\end{comment}

\begin{table}[!h]
\footnotesize
\centering{
\scalebox{0.8}{\begin{tabular}{|c|c|}
\hline
%{} & \multicolumn{1}{|c|}{\textbf{Davis ($\mathbf{pK_{d}}$)}}\\
\textbf{Name of the Models} & {\textbf{Mean Squared Error}}\\
\hline
%\bf{Model} & \bf{MSE}\\
\hline
$1^{st}$ block GCN + 1-D CNN & 0.254\\
\hline
$1^{st}$ block GCN + Bi-LSTM & \bf{0.239}\\
\hline
\end{tabular}
}
}
\caption{Performance comparison of 1-D CNN vs. Bi-LSTM (using \textbf{Davis Dataset ($\mathbf{pK_{d}}$)})}
\label{tab:Ablation1}
\end{table}

\begin{table*}[!th]
\begin{threeparttable}[b]
\centering{
\scalebox{0.7}{\begin{tabular}{|c|c|}
\hline
  \textbf{Name of the Models} & {\textbf{Mean Squared Error}}\\
    \hline
    \hline
    $1^{st}$ block\tnote{1} $\longrightarrow$ $1^{st}$ block + $2^{nd}$ block (\# of layers 2) & 0.237\\
    \hline
    $1^{st}$ block + $2^{nd}$ block$\longrightarrow$ \textbf{DeepGLSTM} & \textbf{0.232}\\
    \hline
    \textbf{DeepGLSTM} $\longrightarrow$ $1^{st}$ block (\# of layers 4) + $2^{nd}$ block (\# of layers 3) + $3^{rd}$ block (\# of layers 2) + $4^{th}$ block (\# of layer 1) & 0.238\\
    \hline
    \end{tabular}
    }
    }
    \caption{Effectiveness of different components of \textbf{DeepGLSTM} (using \textbf{Davis Dataset ($\mathbf{pK_{d}}$)})}
    \label{tab:ablation2}
    \begin{tablenotes}
     \item[1] GCN block
    \end{tablenotes}
    \end{threeparttable}
\end{table*}

\begin{figure}[!t]
   \hspace{-0.49cm}
    \includegraphics[width=0.5\textwidth]{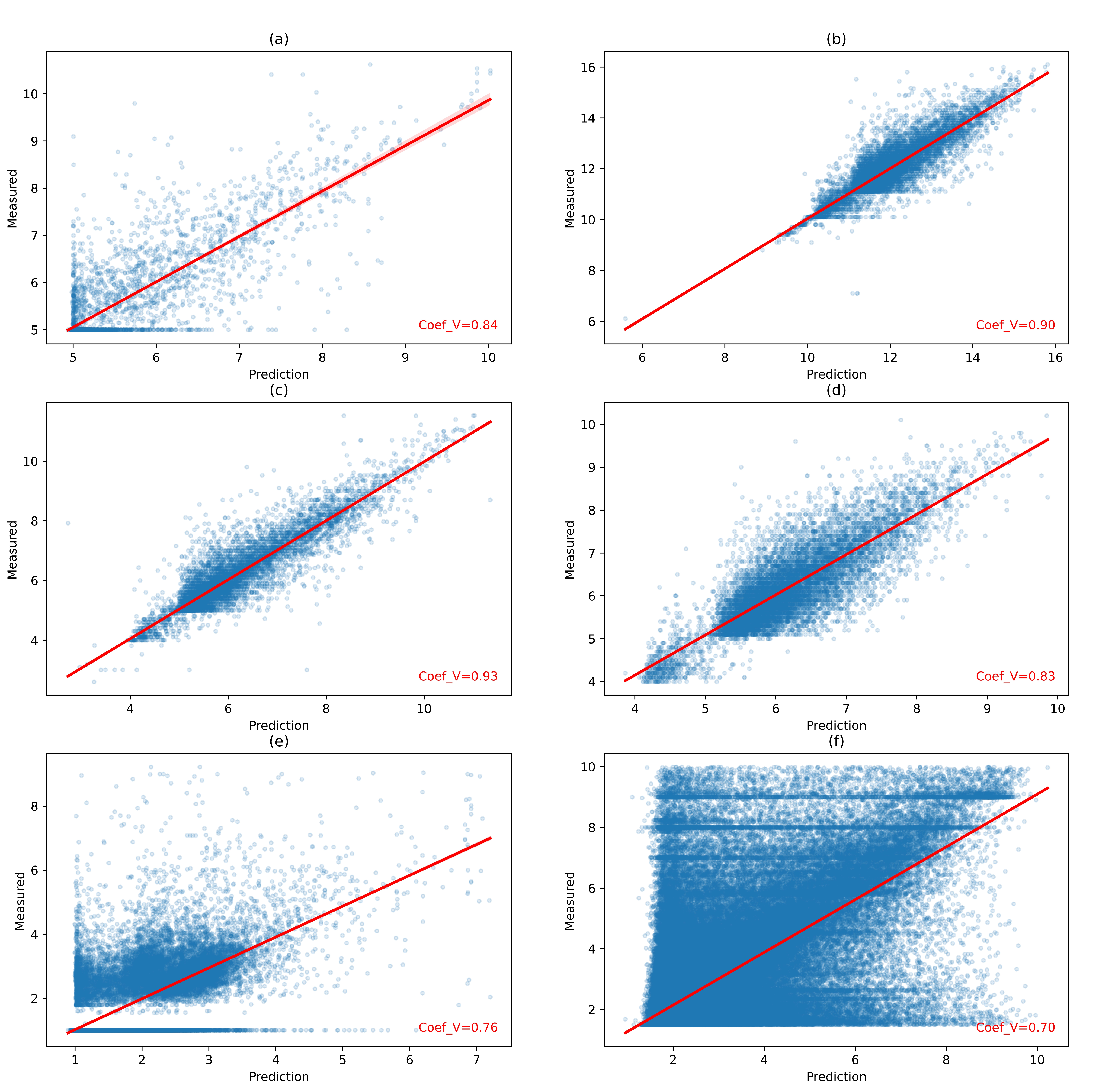}
    \caption{Plots showing \textbf{DeepGLSTM} versus measured binding affinity values for the (a)  Davis dataset ($pK_{d}$) (b) KIBA dataset (KIBA score) (c) DTC dataset ($pK_{i}$) (d) Metz dataset ($pK_{i}$) (e) ToxCast dataset ($AC_{50}$) (f) STITCH dataset (SCORES). In this figure Coef\_V is Pearson correlation score.}
    \label{fig:davis_kiba}
\end{figure}

%\footnotetext{Coef_V is Pearson Correlation Coefficient.}

\begin{comment}

\begin{figure}[!t]
   \hspace{-1cm}
    \includegraphics[scale=0.3]{Updated_fig_new (1).pdf}
    \caption{Plots showing \textbf{DeepGLSTM} versus measured binding affinity values for the (a)  Davis dataset ($pK_{d}$) (b) KIBA dataset (KIBA score) (c) DTC dataset ($pK_{i}$) (d) Metz dataset ($pK_{i}$) (e) ToxCast dataset ($AC_{50}$) (f) STITCH dataset (SCORES). New fig}
    %\label{fig:davis_kiba}
\end{figure}
\end{comment}

\begin{table}[!h]
\footnotesize
\centering{
\scalebox{0.7}{ \begin{tabular}{|c|c|c|}
    \hline
    \textbf{Name of the} & \textbf{Input to} &  \multicolumn{1}{|c|} {\textbf{Mean Squared}}\\
     \textbf{Models} &  \textbf{GCN block} & \textbf{Error}\\
    \hline
    \hline
    $1^{st}$ block GCN  & $A$ & 0.239\\
    \hline
    $2^{nd}$ block GCN  & $A^{2}$ & 0.245\\
    \hline
    $3^{rd}$  block GCN  & $A^{3}$ & 0.244\\
    \hline
    \textbf{DeepGLSTM} & $A$, $A^{2}$, $A^{3}$ & \textbf{0.232}\\
    \hline
    \end{tabular}
    }
    }
    \caption{Effectiveness of using the power graph (using \textbf{Davis Dataset ($\mathbf{pK_{d}}$)})}
    \label{tab:ablation3}
\end{table}

We perform the required computation for \textbf{MSE} (where $P_{i}$ is the predicted binding affinity value and $Y_{i}$ is the original binding affinity value), \textbf{CI} and $\mathbf{r_{m}^{2}}$ \textbf{index} using the equations Eqns. \ref{MSE}, \ref{CI} (where $Z$ is a normalization constant that equals the number of drug-target
pairs with different binding affinity values) and \ref{rm2}, respectively.

\textbf{CI}~\cite{ozturk2018deepdta} measures whether the rank of the predicted binding affinity score and ground-truth scores are same or not. More specifically, when $\delta_i > \delta_j$, a positive score is given iff the predicted score ($b_i$) for the higher affinity $\delta_i$ $>$ the predicted score ($b_j$) for the lower affinity $\delta_j$. Here, $h(x)$ is a step function. 

The metric $\mathbf{r_m^{2}}$~\cite{lin2020deepgs}
is used to evaluate the external prediction performance
of QSAR (Quantitative Structure-Activity Relationship) models. Model's prediction is considered iff $r_m^{2} \geq 0.5$. In Eqn. \ref{rm2}, $r^2$ and $r_0^{2}$ represent the squared correlation coefficient values between the ground-truth and the predicted values with and without intercept, respectively.\\

The metric, C\textsubscript{b}, Combined Score (as mentioned in Eqn.~\ref{comb}) is used to obtain the top-18 (cf. supplementary) FDA-approved drugs for 5 viral proteins of SARS-CoV-2. Since, Davis and KIBA datasets are widely used in literature, we use both of these datasets to implement C\textsubscript{b}. 

Fig. \ref{fig:davis_kiba} shows the predicted value ($p$) versus measured value ($m$) plots on all the datasets. If the predicted value ($p$) closely resembles the measured value ($m$), a model may be considered to be good and therefore, the output values should be close to the red line ($p = m$) as shown in the
figures. From Fig. \ref{fig:davis_kiba}, we can see that our model performs well on Davis, KIBA, DTC and Metz. For ToxCast and STITCH datasets, our model shows performance compared to other state-of-the-art methods. Therefore, we can conclude that the performance of our model is uniformly good over most of the datasets that are available in drug repurposing domain.
We compare performance of our model with the performance of the baseline models mentioned in section \ref{baseline}. The comparative evaluation results are presented in Table~\ref{tab:Model_per1} and Table~\ref{tab:Model_per2}. Our model outperforms all the baseline models in terms of \textbf{CI} on Davis, KIBA, DTC, METZ and ToxCast datasets, in terms of \textbf{MSE} on KIBA, DTC, METZ, ToxCast and STITCH datasets and also in terms of $\mathbf{r_m^{2}}$ on KIBA, DTC and STITCH datasets. Additionally, our model produces relatively comparable results in terms of \textbf{MSE} on Davis dataset and in terms of $\mathbf{r_m^{2}}$ on Davis, Metz and ToxCast datasets. To conclude in a more consolidated way, we use our model to predict drug-target affinity in terms of
C\textsubscript{b} (cf. Eqn.~\ref{comb}) for 2,304 ($n'$) FDA-approved drugs against 5 SARS-CoV-2 viral~\cite{shannon2020remdesivir,jin2020structure,anand2003coronavirus,osipiuk2021structure,ou2020characterization} proteins. As a result, this model yields the required C\textsubscript{b} scores for 11,520 drug-target interactions.

According to the definition, higher C\textsubscript{b} value indicates higher binding affinity between a drug and its corresponding target and vice-versa~\cite{he2017simboost}. So, we first sort the drugs in descending order of their C\textsubscript{b} values predicted by our model. Then we take the top three drugs that bind with receptors and top three drugs that bind with non-receptors. We do this for five highly researched SARS-CoV-2 viral proteins (Glycoprotein S aka Spike protein, RNA-dependent RNA Polymerase aka RdRp, Helicase, Papain-like Proteinase and Open Reading Frame 3a aka ORF3a) and combine them into a set consisting of 18 drugs. The prediction statistics for the selected SARS-CoV-2 viral proteins, in terms of all the binding affinity measuring metrics used, have been provided in the supplementary. Among the drugs that bind with receptors, the ones with the highest C\textsubscript{b} values were, Isoproterenol, Medrysone, Lindane, Mecamylamine, Oxybenzone, Morphine, Estriol, L-Menthol, Ergocalciferol and Methoxyphenamine. Among the drugs that bind with non-receptors, the ones with the highest C\textsubscript{b} values were, Lactulose, Sorbitol, Migalastat, Guaiacol, Cianidanol, Mannitol, Arbutin and Benzoic. Therefore, from the list of drugs with highest binding affinity scores for the viral proteins, researchers may select drugs to conduct further experiments to understand their usage as prospective drugs to treat SARS-CoV-2 patients. As a future work, the receptor based drugs with high binding affinity may be tested for their interactions with important SARS-CoV-2 receptors. Similarly, the non-receptor based drugs with high binding affinity may be tested for their interactions with important enzymes.

\section{Analysis and Discussion}
We carried out ablation study on Davis dataset~\cite{davis2011comprehensive} for our \textbf{DeepGLSTM} model. Initially, we consider GraphDTA~\cite{nguyen2021graphdta} architecture as our baseline model. GraphDTA uses a block similar to the first GCN block of \textbf{DeepGLSTM} for drug compounds and 1-D CNN for protein sequence information. But it is a well-known fact that LSTM performs better than 1\nobreakdash-D CNN to capture the sequential representation \cite{yin2017comparative}. We just replace the 1-D CNN with Bi-LSTM keeping the remaining parts of the GraphDTA architecture same. The performance evaluation of this experiment is reported in Table \ref{tab:Ablation1}. It shows that Bi-LSTM is much more efficient than 1-D CNN.\\   

Additionally, to show the effectiveness of different components of our \textbf{DeepGLSTM} model, we carried out ablation study experiments by incrementally introducing the model components. The result of these ablation study experiments are reported in  Table \ref{tab:ablation2},  which clearly shows the contribution of each component in our model architecture. The results in Table \ref{tab:ablation2} suggest that the third block of GCN with Bi-LSTM contributes the most in the model performance. In most cases, the graph structure of a drug does not consist of many shortest paths that exceed a distance of 3. Therefore, we can conclude that increasing the exponent of a power graph beyond 3 may not lead to any significant change in reachability for most of the nodes, thereby, not contributing significantly towards the effectiveness of the model. The results of Table \ref{tab:ablation2} also indicate that an increase in the exponent of the power graph input, used in the GCN blocks, beyond a certain value may not contribute towards increasing the effectiveness of the model.

To show the effectiveness of power graph representation in our model, we carried out three additional experiments with \textbf{DeepGLSTM}. Firstly, we input the adjacency matrix $A$ of the drug molecules into the first GCN block of \textbf{DeepGLSTM} and the remaining two GCN blocks are removed. Secondly, we pass only the squared representation of $A$ (i.e., $A^{2}$) as input into the second GCN block of \textbf{DeepGLSTM} and the remaining two GCN blocks are removed. Finally, we take the cubic representation of $A$ (i.e., $A^{3}$) as input into the third GCN block of \textbf{DeepGLSTM} and the first two GCN blocks are removed. Thus, using the aforementioned settings we train three different models and note their corresponding \textbf{MSE}s.  The results of these ablation study experiments have been reported in Table \ref{tab:ablation3}. The results show that the second block GCN with input $A^2$ and the third block GCN with input $A^3$ contribute a lot to achieve comparatively low \textbf{MSE} scores. In this way, we introduce \textbf{DeepGLSTM} (i.e., combination of the above three GCN blocks with three different types of adjacency input representations) to  achieve substantially low \textbf{MSE} scores.

\section{Conclusion}
In this paper, we present a novel architecture for the task of predicting binding affinity values between the FDA-approved drugs and viral proteins. Our model uses three blocks of GCN for learning the topological information from the drug molecules. Our Bi-LSTM learns the representation of the protein sequences. We carried out our experiments on the Davis, KIBA, DTC, Metz, ToxCast and STITCH datasets. We also use our model for predicting binding affinity values between 2,304 FDA-approved drugs and 5 viral proteins for SARS-CoV-2. Our model produces state-of-the-art results on the Davis, KIBA, DTC, Metz, ToxCast and STITCH datasets. We have also reported the drugs that show significantly high binding affinity with  some of the most studied viral proteins of SARS-CoV-2.

\section{Acknowledgement}
We thank Siddhartha S Jana for his helpful comments.

\pagebreak
\newpage
\clearpage
\begin{comment}

\begin{table}[]
    \centering
\scalebox{0.8}{    \begin{tabular}{|c|c|c|}
    \hline
         {\bf Pubchem } & {\bf Drug} & {\bf Combined}\\
         {\bf ID} & {\bf Name} & {\bf Score}\\
         & & $\mathbf{pK_{d}, KIBA}$\\
         \hline
         3779 & Isoproterenol & 0.618804939\\
         \hline
         247839 & Medrysone & 0.618199745\\
         \hline
         11333 & Lactulose & 0.617713856\\
         \hline
         176077 & Migalastat & 0.617668157\\
         \hline
         727 & Lindane & 0.617292586\\
         \hline
         1738118 & Aldehydo-N-acetyl-D-glucosamine & 0.617017272\\
         \hline
         16666 & l-Menthol & 0.616839585\\
         \hline
         6251 & Mannitol & 0.616532331\\
         \hline
         5280793 & Ergocalciferol & 0.616387674\\
         \hline
         460 & Guaiacol & 0.61628466\\
         \hline
         162955 & Tixocortol & 0.616254157\\
         \hline
         5288826 & Morphine & 0.615743411\\
         \hline
         5988 & Sucrose & 0.615720423\\
         \hline
         5754 & Hydrocortisone & 0.615651455\\
         \hline
         3082054 & Gaxilose & 0.615615623\\
         \hline
    \end{tabular}}
    \caption{Caption}
    \label{tab:my_label}
\end{table}
\end{comment}

\end{document}